\documentclass[letterpaper, 10 pt, conference]{ieeeconf} 
\IEEEoverridecommandlockouts 
\usepackage[T1]{fontenc}
\usepackage{amsmath,amsfonts,amssymb}
\usepackage{algorithm}
\usepackage{graphicx}
\usepackage{float}
\usepackage{algpseudocode}
\usepackage{pgfplots}
\pgfplotsset{compat=1.17}
\usepgfplotslibrary{groupplots}
\usepackage{array} 
\usepackage{multirow} 
\usepackage{makecell}
\usepackage{xcolor}
\definecolor{reviewgreen}{RGB}{0,128,0}

\newcommand{\reviewnew}[1]{{\color{reviewgreen}#1}}

\usepackage{dblfloatfix} 
\usepackage[font=footnotesize, labelfont=footnotesize]{caption}
\usepackage{subcaption}
\newcolumntype{P}[1]{>{\centering\arraybackslash}p{#1}}
\usepackage{dirtytalk}
\usepackage{booktabs} 
\usepackage{adjustbox}   
\usepackage{microtype}
\usepackage{siunitx}
\usepackage{cite}
\usepackage[hang,flushmargin]{footmisc}
\makeatletter
\let\NAT@parse\undefined
\makeatother
\usepackage[breaklinks,colorlinks]{hyperref}
\renewcommand{\baselinestretch}{0.99}

\renewcommand{\thesubsubsection}{\thesubsection.\alph{subsubsection}}

\usepackage{colortbl}
\definecolor{headercolor}{RGB}{255, 230, 200}   
\definecolor{colhighlight}{RGB}{235, 248, 235}    

\definecolor{csafe}{HTML}{4DB6AC}
\definecolor{cmoderate}{HTML}{FFB74D}
\definecolor{ccritical}{HTML}{FF7043}
\definecolor{cimminent}{HTML}{E53935}

\usepgfplotslibrary{external}
\usepackage{tikz}
\usetikzlibrary{shapes.geometric, arrows.meta, positioning, decorations.pathreplacing}

\renewcommand{\baselinestretch}{0.980}

\begin{document}

\title{\LARGE \bf
Effort-Based Criticality Metrics for Evaluating\\ 3D Perception Errors in Autonomous Driving}

\author{Sharang Kaul$^{1,2}$, Simon Bultmann$^{2}$, Mario Berk$^{1}$ and Abhinav Valada$^{2}$
 \thanks{$^1$CARIAD SE, Germany}%
 \thanks{$^2$Department of Computer Science, University of Freiburg, Germany}%
}

\maketitle
\thispagestyle{empty}
\pagestyle{empty}


\begin{abstract}

Criticality metrics such as time-to-collision (TTC) quantify collision urgency but do not distinguish the operational consequences of false-positive (FP) and false-negative (FN) perception errors. We formulate two error-specific effort metrics: \textit{False Speed Reduction} (FSR), the cumulative velocity loss associated with persistent phantom detections, and \textit{Maximum Deceleration Rate} (MDR), the peak braking demand associated with missed objects under a longitudinal kinematic model. These longitudinal metrics are complemented by \textit{Lateral Evasion Acceleration} (LEA), adapted from prior lateral-evasion kinematics and coupled with reachability-based collision timing. The collision check quantifies the minimum steering effort required to avoid a predicted collision. A dynamically conservative, semantically unfiltered reachability gate selects candidate interactions before frame-level scoring and track-level aggregation. Evaluation on nuScenes and Argoverse~2 shows that \SIrange{65}{93}{\percent} of errors fall below the chosen criticality thresholds. Correlation and threshold analysis indicate that the proposed metrics provide complementary rankings for screening and mining perception failures and are not substitutes for closed-loop safety validation.

\end{abstract}



\section{Introduction}

The safe deployment of automated vehicles requires quantifying perception system errors by both frequency and consequences. Standard detection metrics, such as mean Average Precision (mAP) and the nuScenes Detection Score (NDS), and tracking metrics, such as MOTA and HOTA~\cite{Fong2025nuScenesRP, 9341164}, treat all errors equally. For instance, missing a parked car 50\si{\meter} away in an adjacent lane incurs the same recall penalty as missing a vehicle directly ahead at 10\si{\meter} on a collision course. However, these scenarios require fundamentally different planner responses. This distinction is especially important in open-loop perception evaluation~\cite{lang2022robust,mohan2024progressive}, where detector outputs are compared with ground truth without a downstream planner. Unlike closed-loop testing~\cite{trumpp2023efficient}, where re-planning can compensate for errors, open-loop evaluation provides no feedback to reveal which errors would trigger safety-critical responses.\looseness=-1

Given these challenges, it is essential to develop a principled method to estimate the collision-avoidance effort that a planner would require for each perception error, even without a full control stack. By introducing effort-based metrics for open-loop perception testing, we can capture the criticality of errors early in development. This, in turn, enables safety-aware comparison of perception pipelines and trackers on standardized benchmarks. Criticality metrics such as TTC, Time-to-Brake (TTB), Time Headway (THW), and Deceleration to Safety Time (DST)~\cite{Westhofen_2022} measure the urgency of collisions. However, they do not capture the consequences of each type of perception error. Applied to perception outputs, these metrics become ambiguous. For example, a false positive (FP) with low TTC leads to unnecessary braking, not real danger. In contrast, a false negative (FN) increases braking demand once detected. DST primarily focuses on braking effort but serves as a ground-truth metric. It does not distinguish between FP-induced phantom braking and FN-induced emergency braking. To our knowledge, no prior metric directly quantifies the effort required to avoid collisions for each ego vehicle perception error.

To address this limitation, we introduce two complementary metrics grounded in longitudinal vehicle kinematics: (i) \textit{False Speed Reduction (FSR)} accumulates the unwarranted velocity loss associated with a persistent FP, and (ii) \textit{Maximum Deceleration Rate (MDR)} records the peak braking demand associated with an FN. We complement them with \textit{Lateral Evasion Acceleration (LEA, $a_{\perp}$)}, adapted from~\cite{9808147} and coupled with reachability-based collision timing from~\cite{schneider2021towards}. All three metrics are evaluated after a reachability-based gate check~\cite{schneider2021towards, althoff2010computing}. The gate tests the dynamic possibility of collision under bounded motion. It does not encode lanes, road boundaries, or traffic rules.

The main contributions of this work are:
(i) \textit{Error-specific effort aggregation:} FSR and MDR distinguish persistent FP-associated braking cost from peak FN-associated braking demand, complementing TTC, DRAC, DST, and TET rather than replacing them.
(ii) \textit{Modular candidate collision check:} Ellipsoidal reachable sets, frame-level matching, and track-level aggregation are combined in an offline benchmark pipeline. The proposed conservative collision gate can be replaced by a tighter geometric or map-aware test.
(iii) \textit{Longitudinal-lateral analysis:} LEA provides a lateral complement to FSR and MDR, exposing braking-steering trade-offs that a single longitudinal score cannot represent.


\begin{figure*}[t]
\centering
\includegraphics[width=0.9\textwidth]{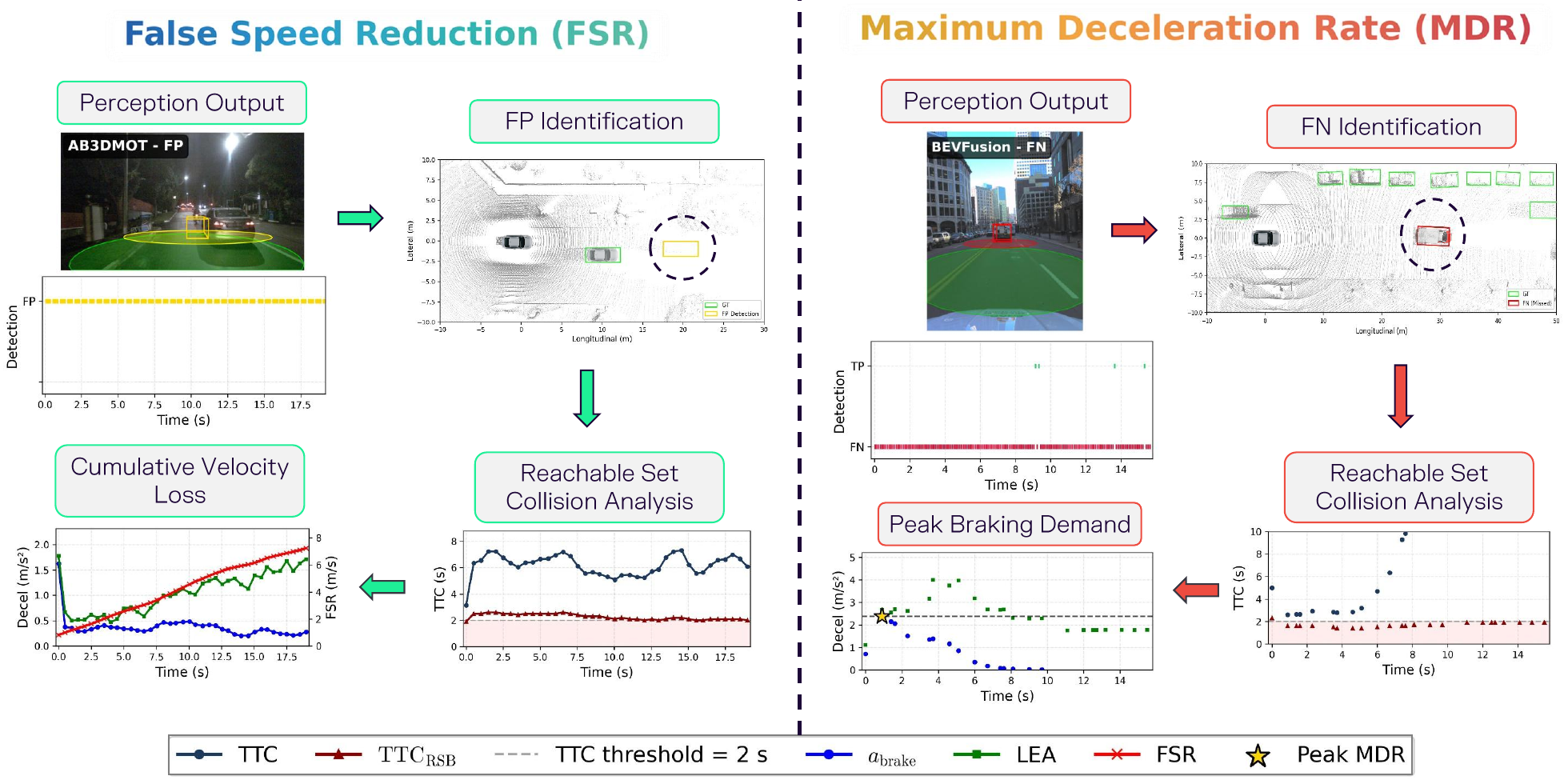}
\caption{Effort-based criticality assessment of perception errors. The pipeline identifies false positives (FPs) and false negatives (FNs) by matching tracker outputs against ground-truth (GT) annotations, performs reachable-set collision analysis, and computes effort metrics. Regular GT boxes are shown in green. A missed GT box is recoloured red to indicate its FN status. \textit{False Speed Reduction} (FSR) captures the cumulative velocity loss from persistent FPs, and \textit{Maximum Deceleration Rate} (MDR) captures the peak braking demand from missed objects. \textbf{Left side:} FP from AB3DMOT on nuScenes where a phantom detection (yellow) persists for $19$\,\si{\second} ($39$ frames) at ${\sim}20$\,\si{\meter} ahead with no matching GT object. FSR accumulates to $7.2$\,\si{\meter\per\second}. \textbf{Right side:} FN from BEVFusion on Argoverse~2. The red box is the GT annotation of a real vehicle in the ego lane ($R_{\text{lat}}\!\approx\!0.3$\,\si{\meter}). It is an FN because no BEVFusion prediction is matched to this GT object for ${\sim}16$\,\si{\second}. MDR peaks at $2.4$\,\si{\meter\per\second\squared}. LiDAR bird's-eye-view panels show multi-sweep point clouds in gray. Metrics are computed only during frames with perception errors.}
\label{fig:combined_motivation}
\vspace{-0.5cm}
\end{figure*}

\section{Related Work}
\label{sec:related_work}

{\parskip=0pt
\noindent\textit{Safety-aware Perception Evaluation}: 
Recent works weigh detections by safety relevance. Wolf~\textit{et~al.}~\cite{9564734} weight pedestrian detections by position and estimated TTC. Volk~\textit{et~al.}~\cite{9294708} propose a safety metric that incorporates object relevance relative to the observer. Ceccarelli~\textit{et~al.}~\cite{Ceccarelli_2023} introduce an Object Criticality Model (OCM) based on distance, trajectory alignment, and TTC, and later show that criticality filtering improves trajectory planning quality~\cite{9157486}. Although these methods distinguish important misses from benign ones, they remain \emph{binary} (detected vs.\ missed) and do not quantify the magnitude of the resulting braking or evasion effort.}

{\parskip=2pt
\noindent\textit{Ego-centric Evaluation}: 
USC~\cite{10919937} requires that a prediction fully covers its ground-truth object from the ego's viewpoint perspective and BEV, with higher coverage correlating with lower collision danger. EC-IoU~\cite{10801740} weights IoU toward the closer portion of each box, improving mAP when used as a training loss. Philion~\textit{et~al.}~\cite{9157486} bypass detection metrics entirely, evaluating perception through a planner-centric metric (PKL) that measures trajectory divergence. These formulations tie evaluation closer to the driving task but do not yield interpretable physical quantities for individual error events.}


{\parskip=2pt
\noindent\textit{Safety-aware Training}: 
Lyssenko~\textit{et~al.} incorporate safety into pedestrian detection through reachability-based TTC~\cite{9981309}, a credibility metric for non-credible FNs~\cite{Lyssenko2024AFC}, and a safety-adapted loss~\cite{Lyssenko2024ASL}. Cheng~\textit{et~al.}~\cite{Cheng2020SafetyAwareHO} harden 3D detectors by focusing on critical zones. These approaches embed safety awareness into the model but target specific classes and do not yield a generalized scalar metric for the avoidance effort induced by each FP or FN. In contrast, FSR and MDR translate each error into a physical quantity: cumulative speed reduction (\si{\meter\per\second}) for FPs, peak braking deceleration (\si{\meter\per\second\squared}) for FNs, and are class-agnostic, requiring no planning stack.}

{\parskip=2pt
\noindent\textit{Reachability Analysis for Collision Prediction}: 
Rather than propagating one constant-velocity trajectory, reachability analysis encloses future states under bounded dynamics. Althoff~\textit{et~al.}~\cite{althoff2010computing} develop set-based methods for hybrid systems, and Schneider~\textit{et~al.}~\cite{schneider2021towards} apply computationally practical ellipsoidal over-approximations to driving. Philipp~\textit{et~al.}~\cite{9808147} provide a kinematic model for lateral evasion acceleration. Within the assumed acceleration bounds, ellipsoidal over-approximations are dynamically conservative. We adopt this combination for bounded-motion coverage and computational tractability. The gating filter can be easily tightened or replaced with a stricter collision test, such as an oriented bounding-box overlap check or a map-aware lane-relevance gate, without changing the metric formulation.}


\section{Technical Approach}

We derive effort-based criticality metrics that quantify the collision-avoidance demand induced by perception errors. We derive two complementary metrics from one-dimensional longitudinal kinematics: \textit{False Speed Reduction} (FSR) quantifies unnecessary braking due to persistent FPs, and \textit{Maximum Deceleration Rate} (MDR) quantifies the peak evasive braking demand arising from missed detections. In both cases, we compute the constant deceleration required to avoid a collision, given reaction time and relative motion. The per-frame estimates are then aggregated into scalar criticality measures.

\subsection{Motivation for FSR and MDR}\label{Motivation_FSR_MDR}
Classical TTC, given by $R/\Delta v$, uses the longitudinal separation $R$ and closing speed $\Delta v$ under a constant-velocity assumption. It can therefore underestimate collision risk when agents accelerate or brake. We complement it with the reachability-based collision test described in Sec.~\ref{sec:reachability}. We define $\mathrm{TTC_{RSB}}$ as the earliest time at which the ego and object reachable sets overlap. This overlap can occur earlier than the constant-velocity TTC. All effort metrics are computed only during perception-error frames.

\textit{FP Scenario:} A persistent misdetection from AB3DMOT is tracked for $19$\,\si{\second} ($39$ frames) at ${\sim}25$\,\si{\meter} ahead of the ego vehicle ($v_{\text{ego}} \approx 8$\,\si{\meter\per\second}) in Fig.~\ref{fig:combined_motivation} (left). Classical TTC ranges between $3$\,\si{\second} and $7$\,\si{\second}, confirming a genuine closing geometry. $\mathrm{TTC_{RSB}}$ drops to ${\sim}2.7$\,\si{\second} when the reachable sets overlap. The required per-frame braking magnitude $a_{\text{brake}}$ remains moderate between $0.2$ and $1.6$\,\si{\meter\per\second\squared}, yet FSR accumulates to $7.2$\,\si{\meter\per\second} (${\sim}26$\,\si{\kilo\meter\per\hour}), quantifying the sustained operational cost invisible to any single-frame metric.\looseness=-1

\textit{FN Scenario:} On Argoverse~2, the red box in Fig.~\ref{fig:combined_motivation} (right) marks the GT annotation of a real vehicle with a lateral offset $R_{\text{lat}} \approx 0.3$\,\si{\meter} from the ego path. The object is classified as an FN because no BEVFusion prediction is matched to it for ${\sim}16$\,\si{\second}. While closing ($t{=}0$-$5$\,\si{\second}), $a_{\text{brake}}$ peaks at MDR\,$=2.4$\,\si{\meter\per\second\squared}. As both vehicles decelerate and eventually stop, $\Delta v{\to}0$ and $a_{\text{brake}}{\to}0$. Throughout, $\mathrm{TTC_{RSB}}$ remains near $1$-$2$\,\si{\second} because the bounded-acceleration sets continue to expand. LEA ranges from $1.1$ to $4.0$\,\si{\meter\per\second\squared} as the available evasion time changes with the predicted collision time. The two metrics thus decouple: MDR is driven by the closing kinematics given by Eqs.~\ref{eq:mdr_base}-\ref{eq:mdr_peak} and correctly decays as the threat diminishes, whereas LEA is driven by the reachable-set collision time and remains elevated even after both vehicles have stopped, reflecting conservatism inherited from the filter.\looseness=-1

\subsection{Metric Design: FSR and MDR}\label{Metric_design}

All longitudinal quantities are projected onto the current ego-heading axis. Let $v_{\text{ego}}\geq0$ denote ego speed, $t_{\text{react}}$ the sensor-to-actuator delay, and $R>0$ the object-center range ahead of the ego. For FPs, $v_{\text{FP}}$ and $R_{\text{FP}}$ are the perceived longitudinal speed and range. For FNs, $v_{\text{FN}}$, $a_{\text{FN}}$, and $R_{\text{FN}}$ are the annotated longitudinal speed, estimated longitudinal acceleration, and range. We use $a_{\text{brake}}\geq0$ throughout as a braking magnitude, not as a signed acceleration. Its effect on signed ego velocity and displacement is therefore represented by $-a_{\text{brake}}t$ and $-\tfrac{1}{2}a_{\text{brake}}t^2$.

\subsubsection{False Speed Reduction (FSR) for Persistent FPs} For each frame in which the FP persists, the deceleration $a_{\text{brake,FP},i}$ is derived by equating the distance covered by the ego vehicle to the FP's distance plus separation, by the time speeds match. The per-frame kinematic constraint is:
\begin{equation} \label{eq:fsr_base} \begin{aligned}
&
v_{\text{ego}} \times t_{\text{react}} + \frac{v_{\text{ego}}(v_{\text{ego}} - v_{\text{FP}})}{a_{\text{brake,FP}}} - \frac{1}{2}\frac{(v_{\text{ego}} - v_{\text{FP}})^2}{a_{\text{brake,FP}}} \\
&\quad - R_{\text{FP}} - v_{\text{FP}} \times t_{\text{react}} - \frac{v_{\text{FP}}(v_{\text{ego}} - v_{\text{FP}})}{a_{\text{brake,FP}}} = 0
\end{aligned}
\end{equation}

The instantaneous deceleration can be solved for (assuming $v_{\text{ego}} > v_{\text{FP}}$ and $R_{\text{FP}} > (v_{\text{ego}}-v_{\text{FP}}) \times t_{\text{react}}$):
\begin{equation} \label{eq:fsr_simplified}
a_{\text{brake,FP},i} = \frac{(v_{\text{ego}}-v_{\text{FP}})^{2}}{2(R_{\text{FP}}-(v_{\text{ego}}-v_{\text{FP}}) \times t_{\text{react}})}
\end{equation}
The average required deceleration over the event is then:
\begin{equation}
a_{\text{avg}} = \frac{\sum_{i=1}^{N_{\text{frames}}} a_{\text{brake,FP},i}}{N_{\text{frames}}},
\end{equation}
where $N_{\text{frames}}$ is the number of consecutive error frames. In the implementation, $a_{\text{avg}}$ is computed over the frames with positive braking demand ($a_{\text{brake,FP},i}>0$); frames in which the collision gate predicts no overlap contribute no demand but still count toward the error duration. The final FSR metric is the total error duration in seconds multiplied by the average deceleration, providing a measure of the total unwarranted braking impulse:
\begin{equation} \label{eq:fsr_cumulative}
\text{FSR} = (N_{\text{frames}} \times T_{\text{cycle}}) \times a_{\text{avg}},
\end{equation}
where $T_{\text{cycle}}$ is the time between successive annotation frames (e.g.\ $0.5$\,\si{\second} at $2$\,\si{\hertz} for nuScenes, $0.1$\,\si{\second} at $10$\,\si{\hertz} for Argoverse~2). This cumulative formulation penalizes persistent FPs more heavily than transient ones.


\subsubsection{Maximum Deceleration Rate (MDR) for Critical FNs}
For an FN frame, define the closing speed after the reaction interval as
$\Delta v_r=v_{\text{ego}}-v_{\text{FN}}-a_{\text{FN}}t_{\text{react}}$. Here, $a_{\text{FN}}$ is signed: for a FN, $a_{\text{FN}}>0$ denotes forward acceleration and $a_{\text{FN}}<0$ denotes deceleration. In contrast, $a_{\text{brake,FN}}\geq0$ is the magnitude of ego braking. Consequently, after braking begins, relative acceleration is $-(a_{\text{brake,FN}}+a_{\text{FN}})$. For $\Delta v_r>0$, a finite positive time to eliminate the closing speed requires $a_{\text{brake,FN}}+a_{\text{FN}}>0$. The corresponding time is
\begin{equation}
t_b=\frac{\Delta v_r}{a_{\text{brake,FN}}+a_{\text{FN}}},
\quad \Delta v_r>0,\quad t_b>0.
\label{eq:mdr_base}
\end{equation}
Equating ego and object positions at $t_{\text{react}}+t_b$ gives
\begin{equation}
\label{eq:mdr_base_review}
\begin{aligned}
&v_{\text{ego}}t_{\text{react}}+v_{\text{ego}}t_b
-\tfrac{1}{2}a_{\text{brake,FN}}t_b^2\\
&\quad=R_{\text{FN}}+v_{\text{FN}}(t_{\text{react}}+t_b)
+\tfrac{1}{2}a_{\text{FN}}(t_{\text{react}}+t_b)^2 .
\end{aligned}
\end{equation}
These equations are solved numerically for $a_{\text{brake,FN}}\geq0$.
Frames behind the ego, and non-closing frames with a non-decelerating object,
receive zero demand. Solutions with non-positive $t_b$ are rejected. A
non-positive remaining gap after reaction or a required value above the
vehicle limit is reported at the \SI{10}{\meter\per\second\squared} cap. The MDR is the per-frame maximum deceleration during the FN event:
\begin{equation} \label{eq:mdr_peak}
\text{MDR} = \max_{i \in {\text{frames where FN exists}}} (a_{\text{brake,FN},i})
\end{equation}

A high MDR indicates that, at some point during the miss, a very severe braking manoeuvre would have been necessary. Unlike FSR, which accumulates over the event duration, MDR captures the single most dangerous instant. Both metrics yield a scalar per error track.

\begin{figure}[t]
\centering
\includegraphics[width=\columnwidth]{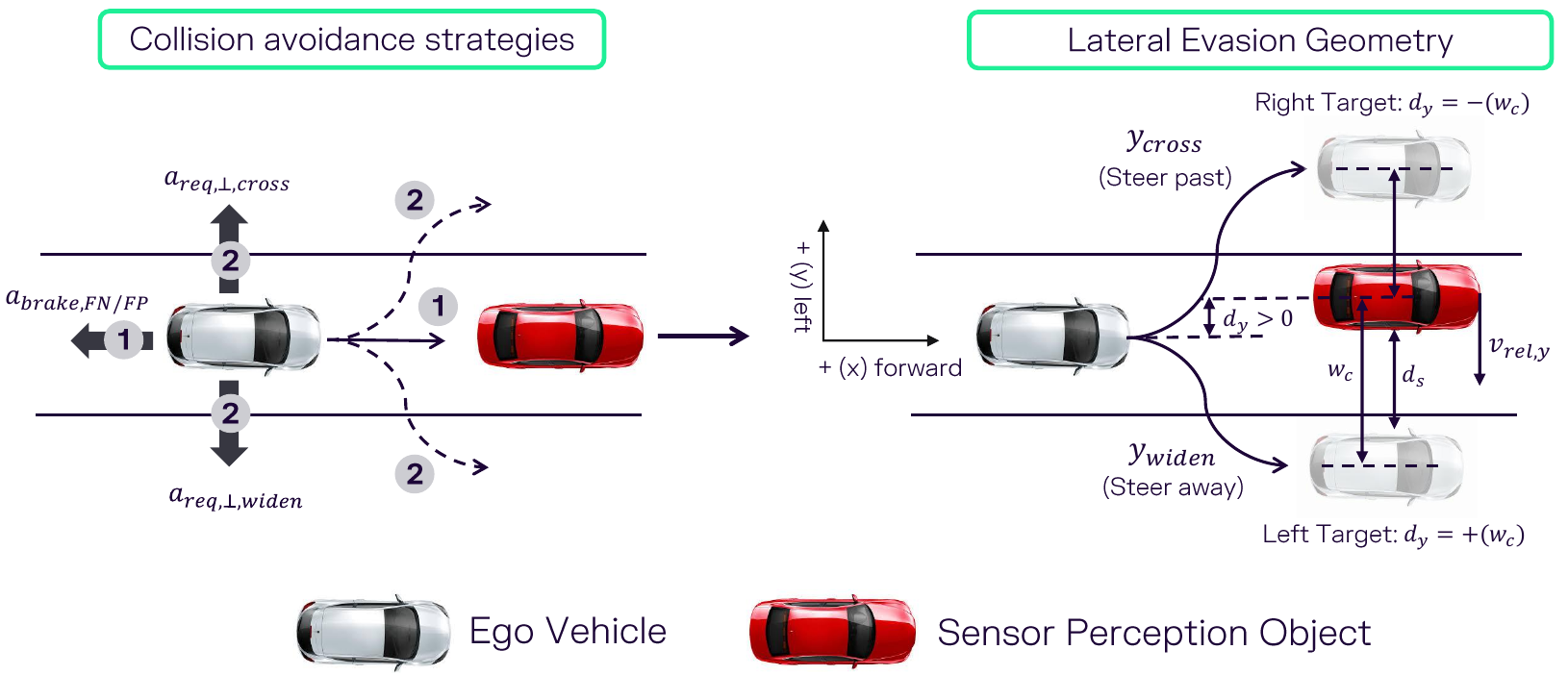}
\caption{Collision avoidance strategies between the ego vehicle (gray) and a perception error (red). \textbf{Left:} longitudinal braking ($a_{\text{brake}}$) and lateral evasion are evaluated independently. \textbf{Right:} lateral-evasion geometry. The signed candidates target relative offsets $d_y=+w_c$ and $d_y=-w_c$ under the ego-frame convention $+x$ forward and $+y$ left.}
\label{fig:technical_approach}
\vspace{-0.5cm}
\end{figure}

\subsection{Lateral Evasion Acceleration (LEA)}\label{sec:lateral}

While longitudinal metrics capture braking demand, collision avoidance often involves steering as shown in Fig.~\ref{fig:technical_approach}. We compute the minimum lateral acceleration $(a_{\text{req},\perp})$ required to evade a predicted collision, triggered whenever the reachable set identifies a future overlap at time $t_{\text{coll}}$. Following the kinematic framework in~\cite{9808147}, we evaluate two distinct evasion strategies: \say{widening} the current lateral gap or \say{crossing} to the opposite side. Let $d_{y}$ denote the current signed lateral offset between vehicle centers, and $w_{\text{c}}$ be the minimum lateral separation required for clearance:
\begin{equation}
w_{\text{c}} = \frac{W_{\text{ego}} + W_{\text{obj}}}{2} + d_{\text{s}},
\end{equation}
where $d_{\text{s}}$ ensures a safety margin.

We use the ego-frame convention shown
in Fig.~\ref{fig:technical_approach}: $+x$ points forward and $+y$ points to
the left of the ego vehicle. The lateral offset and relative velocity are
$d_y=y_{\text{obj}}-y_{\text{ego}}$ and
$v_{\text{rel},y}=v_{\text{obj},y}-v_{\text{ego},y}$, respectively. Thus, an
object to the left of the ego has $d_y>0$, while an object to its right has
$d_y<0$. Over
the evasion window $T_{\text{eva}}=t_{\text{coll}}-t_{\text{react}}$, the
offset without active steering is
\begin{equation}
d_y^{\text{pred}}=d_y+v_{\text{rel},y}T_{\text{eva}}.
\label{eq:dy_pred_review}
\end{equation}
The two clearance targets are $d_y=+w_{\text{c}}$ and $d_y=-w_{\text{c}}$.
Because ego displacement changes the relative offset with the opposite sign,
the required signed ego displacements are
\begin{align}
\Delta y_{+} &= d_y^{\text{pred}}-w_{\text{c}}, \label{eq:y_plus_review}\\
\Delta y_{-} &= d_y^{\text{pred}}+w_{\text{c}}. \label{eq:y_minus_review}
\end{align}
Their signs retain the steering direction. The corresponding signed lateral
accelerations are
\begin{equation}
a_{y,s}=\frac{2\Delta y_s}{T_{\text{eva}}^2},
\qquad s\in\{+,-\}.
\label{eq:ay_signed_review}
\end{equation}
If $(d_y^{\text{pred}})^2\geq w_{\text{c}}^2$, passive motion already provides
the required clearance and no lateral acceleration is needed. Otherwise, we
select
\begin{equation}
s^*=\arg\min_{s\in\{+,-\}}|a_{y,s}|
\label{eq:lea_signed_review}
\end{equation}
\begin{equation}
a_{\text{req},\perp}=a_{y,s^*},\quad
\mathrm{LEA}=|a_{\text{req},\perp}|
\label{eq:lea_final}
\end{equation}
The sign of $a_{\text{req},\perp}$ indicates the steering direction, while LEA
reports its magnitude.

\textit{Motion Model:}\label{sec:motionmodel}
FPs use constant velocity because a phantom track has no defensible physical acceleration. FNs use piecewise constant longitudinal acceleration over the short avoidance interval, while relative lateral motion is constant velocity~\cite{9808147, Westhofen_2022}. These assumptions favor interpretability and reproducibility over prediction fidelity. The interface is modular: a learned or set-valued predictor can replace these roll-outs while preserving the FSR/MDR track aggregation.

\subsection{Reachability-Based Collision Filtering}\label{sec:reachability}

The longitudinal metrics in Sec.~\ref{Metric_design} do not determine whether an object lies on the ego path. We therefore use reachability analysis~\cite{schneider2021towards, althoff2010computing} to filter object-ego pairs before computing effort. This test asks whether an overlap is possible under the assumed motion bounds. At prediction time $\tau$, each vehicle is represented by an ellipsoid $\mathcal{R}(\tau)$. Each ellipsoid is centered on its constant-velocity prediction and inflated symmetrically according to the admissible acceleration range $a_{\text{lon}}\in[a_{\min},a_{\max}]$:
\begin{align}
c_{\text{lon}}(\tau)
&=p_{0,\text{lon}}+v_{0,\text{lon}}\tau\\
\sigma_{\text{lon}}(\tau)
&=\frac{L}{2}+\frac{a_\text{lon,max}-a_\text{lon,min}}{4}\tau^2.
\end{align}
For the bounds used here, $[a_{\min},a_{\max}]=[-3,2]$\,\si{\meter\per\second\squared}. Centering each ellipsoid on the constant-velocity trajectory keeps the check agnostic to the sign of the acceleration deviation, and the symmetric growth over-approximates the forward closing extent that governs lead-vehicle collisions. Laterally, we use symmetric bounds, giving $c_{\text{lat}}(\tau)=p_{0,\text{lat}}+v_{0,\text{lat}}\tau$ and $\sigma_{\text{lat}}(\tau)=W/2+\tfrac{1}{2}a_{\text{lat,max}}\tau^2$. The collision time is the first sampled time at which the ego and object ellipsoids overlap. We sample at $\Delta t=0.1$\,\si{\second} up to $T_{\text{horizon}}=5.0$\,\si{\second}. Only pairs that overlap within this horizon are scored.

\textit{SAT-based Alternative Gate:}
We also implement a tighter \emph{Separating Axis Theorem} (SAT) collision gate~\cite{10.1145/3727353.3727481, 9808147}. Both vehicles are propagated as oriented bounding boxes under one kinematic rollout, and overlap is tested on the four candidate separating axes. SAT replaces a reachable volume with one predicted footprint, admitting fewer tracks but potentially missing collisions enabled by future manoeuvres. Like RSB, SAT contains no lane or map semantics. Therefore, "tighter" does not imply a semantic false-positive guarantee.

\begin{table}[t]
\centering
\caption{Parameters for reachability-based collision filtering and criticality computation}
\label{tab:reach_params}
\setlength{\tabcolsep}{3pt}
\begin{tabular}{llc|llc}
\toprule
\textbf{Parameter} & \textbf{Symbol} & \textbf{Value} & \textbf{Parameter} & \textbf{Symbol} & \textbf{Value} \\
\midrule
\multicolumn{3}{l|}{\underline{\textit{Reachable set bounds}}} & \multicolumn{3}{l}{\underline{\textit{Capability limits}}} \\
Forward acc. & $a_{\text{lon,max}}$ & \num{2.0} & Max braking & $a_{\text{brake,max}}$ & \num{10.0} \\
Braking & $a_{\text{lon,min}}$ & $-$\num{3.0} & Max lateral acc. & $a_{\text{lat,cap}}$ & \num{5.0} \\
Lateral acc. & $a_{\text{lat,max}}$  & \num{2.0} & & & \\
\midrule
\multicolumn{3}{l|}{\underline{\textit{Vehicle \& scenario}}} & \multicolumn{3}{l}{\underline{\textit{Timing}}} \\
Length & $L$ & \num{4.5}\,\si{\meter} & Reaction time  & $t_{\text{react}}$ & \num{0.3}\si{\second} \\
Width & $W$ & 1.8\si{\meter} & Horizon & $T_{\text{hor}}$ & \num{5.0}\si{\second} \\
Safety margin & $d_s$ & \num{0.5}\si{\meter} & Time step & $\Delta t$  & 0.1\si{\second} \\
\bottomrule
\end{tabular}
\begin{flushleft}
\scriptsize All accelerations and braking values in \si{\meter\per\second\squared}. \textit{Reachable set bounds} represent typical driving behaviour and govern the growth rate of the reachable set for collision filtering (Sec.~\ref{sec:reachability}). They are intentionally smaller than the \textit{capability limits}, which represent physical vehicle limits used to cap computed FSR, MDR, and LEA values (Sec.~\ref{Metric_design}). Using capability-level accelerations (e.g.\ $10$\,\si{\meter\per\second\squared}) for reachability prediction would yield excessively large reachable sets, flagging nearly all objects as collision risks.
\end{flushleft}
\vspace{-0.6cm}
\end{table}

\begin{table*}[t]
\centering
\caption{Safety-critical evaluation of object detection models with track-level aggregation on nuScenes and Argoverse~2 validation sets. \textbf{Bold} represents best value per dataset-class group, and \underline{underline} represents second best.}
\label{tab:comprehensive_results}
\scriptsize
\setlength{\tabcolsep}{2.5pt}
\begin{tabular}{ll cc rr rr r >{\columncolor{colhighlight}}r >{\columncolor{colhighlight}}r >{\columncolor{colhighlight}}r >{\columncolor{colhighlight}}r >{\columncolor{colhighlight}}r >{\columncolor{colhighlight}}r >{\columncolor{colhighlight}}r >{\columncolor{colhighlight}}r >{\columncolor{colhighlight}}r rr}
\toprule
& & & & \multicolumn{2}{c}{\textbf{Track Errors}} & \multicolumn{2}{c}{\textbf{Critical}} & & \multicolumn{3}{c}{\cellcolor{colhighlight}\textbf{Mean Effort}} & \multicolumn{3}{c}{\cellcolor{colhighlight}\textbf{Cum.\ Effort}} & \multicolumn{3}{c}{\cellcolor{colhighlight}\textbf{Worst Event}} & \multicolumn{2}{c}{\textbf{Detection}} \\
\cmidrule(lr){5-6} \cmidrule(lr){7-8} \cmidrule(lr){10-12} \cmidrule(lr){13-15} \cmidrule(lr){16-18} \cmidrule(lr){19-20}
\textbf{Dataset} & \textbf{Pipeline} & \textbf{Mod.} & \textbf{Class} & \textbf{FN} & \textbf{FP} & \textbf{FN} & \textbf{FP} & \textbf{TC} & \textbf{MDR} & \textbf{FSR} & \textbf{LEA} & \textbf{MDR} & \textbf{FSR} & \textbf{LEA} & \textbf{MDR} & \textbf{FSR} & \textbf{LEA} & \textbf{Prec.} & \textbf{Rec.} \\
\midrule
\multirow{6}{*}{nuScenes}
 & \multirow{2}{*}{AB3DMOT} & \multirow{2}{*}{L} & Car & \textbf{\num{3252}} & \num{36239} & \textbf{\num{696}} & \num{4816} & \num{5632} & \textbf{\num{2.32}} & \num{1.66} & \underline{\num{0.05}} & \textbf{\num{7550}} & \num{60242} & \num{2061} & \num{10.0} & \num{200.0} & \num{5.0} & \num{0.32} & \textbf{\num{0.70}} \\
 & & & Truck & \textbf{\num{644}} & \num{23262} & \textbf{\num{163}} & \num{2616} & \num{2799} & \textbf{\num{2.73}} & \num{1.39} & \textbf{\num{0.04}} & \textbf{\num{1759}} & \num{32247} & \num{872} & \num{10.0} & \num{165.0} & \num{5.0} & \num{0.11} & \textbf{\num{0.57}} \\
\cmidrule(lr){2-20}
 & \multirow{2}{*}{CenterPoint} & \multirow{2}{*}{L} & Car & \underline{\num{3554}} & \num{40172} & \underline{\num{764}} & \underline{\num{3468}} & \underline{\num{5101}} & \textbf{\num{2.32}} & \textbf{\num{0.99}} & \textbf{\num{0.04}} & \underline{\num{8247}} & \underline{\num{39955}} & \underline{\num{1670}} & \num{10.0} & \underline{\num{155.0}} & \num{5.0} & \underline{\num{0.41}} & \underline{\num{0.67}} \\
 & & & Truck & \underline{\num{654}} & \underline{\num{12740}} & \underline{\num{182}} & \underline{\num{1112}} & \underline{\num{1410}} & \underline{\num{2.93}} & \underline{\num{1.01}} & \underline{\num{0.04}} & \underline{\num{1915}} & \underline{\num{12873}} & \underline{\num{503}} & \num{10.0} & \underline{\num{80.0}} & \num{5.0} & \underline{\num{0.24}} & \underline{\num{0.55}} \\
\cmidrule(lr){2-20}
 & \multirow{2}{*}{BEVFusion} & \multirow{2}{*}{L+C} & Car & \num{3865} & \textbf{\num{3440}} & \num{919} & \textbf{\num{454}} & \textbf{\num{1372}} & \num{2.53} & \underline{\num{1.37}} & \num{0.08} & \num{9766} & \textbf{\num{4723}} & \textbf{\num{599}} & \num{10.0} & \textbf{\num{49.9}} & \num{5.0} & \textbf{\num{0.86}} & \num{0.60} \\
 & & & Truck & \num{687} & \textbf{\num{2197}} & \num{222} & \textbf{\num{96}} & \textbf{\num{285}} & \num{3.39} & \textbf{\num{0.59}} & \num{0.05} & \num{2330} & \textbf{\num{1297}} & \textbf{\num{145}} & \num{10.0} & \textbf{\num{60.0}} & \num{5.0} & \textbf{\num{0.74}} & \num{0.43} \\
\midrule
\multirow{6}{*}{Argoverse~2}
 & \multirow{2}{*}{AB3DMOT} & \multirow{2}{*}{L} & Car & \num{9074} & \underline{\num{56614}} & \num{2024} & \underline{\num{312}} & \underline{\num{3312}} & \num{2.34} & \textbf{\num{0.09}} & \underline{\num{0.07}} & \num{21214} & \underline{\num{5315}} & \underline{\num{4336}} & \num{10.0} & \underline{\num{14.0}} & \num{5.0} & \underline{\num{0.46}} & \num{0.54} \\
 & & & Truck & \underline{\num{560}} & \underline{\num{4526}} & \underline{\num{140}} & \underline{\num{46}} & \underline{\num{536}} & \underline{\num{2.65}} & \textbf{\num{0.13}} & \underline{\num{0.22}} & \underline{\num{1484}} & \underline{\num{588}} & \underline{\num{1139}} & \num{10.0} & \textbf{\num{8.0}} & \num{5.0} & \underline{\num{0.34}} & \underline{\num{0.44}} \\
\cmidrule(lr){2-20}
  & \multirow{2}{*}{CenterPoint} & \multirow{2}{*}{L} & Car & \textbf{\num{8162}} & \num{108054} & \textbf{\num{1560}} & \num{372} & \num{4488} & \textbf{\num{2.01}} & \underline{\num{0.11}} & \textbf{\num{0.04}} & \textbf{\num{16419}} & \num{12025} & \num{5110} & \num{10.0} & \num{21.0} & \num{5.0} & \num{0.34} & \textbf{\num{0.60}} \\
 & & & Truck & \textbf{\num{548}} & \num{9126} & \textbf{\num{128}} & \num{85} & \num{823} & \textbf{\num{2.55}} & \underline{\num{0.18}} & \textbf{\num{0.16}} & \textbf{\num{1400}} & \num{1613} & \num{1591} & \num{10.0} & \num{16.0} & \num{5.0} & \num{0.24} & \textbf{\num{0.47}} \\
\cmidrule(lr){2-20}
 & \multirow{2}{*}{BEVFusion} & \multirow{2}{*}{L+C} & Car & \underline{\num{8617}} & \textbf{\num{7490}} & \underline{\num{1827}} & \textbf{\num{100}} & \textbf{\num{1123}} & \underline{\num{2.24}} & \num{0.24} & \num{0.08} & \underline{\num{19261}} & \textbf{\num{1780}} & \textbf{\num{1331}} & \num{10.0} & \textbf{\num{9.9}} & \num{5.0} & \textbf{\num{0.86}} & \underline{\num{0.56}} \\
 & & & Truck & \num{597} & \textbf{\num{747}} & \num{178} & \textbf{\num{16}} & \textbf{\num{205}} & \num{3.11} & \num{0.30} & \num{0.41} & \num{1856} & \textbf{\num{222}} & \textbf{\num{548}} & \num{10.0} & \underline{\num{8.3}} & \num{5.0} & \textbf{\num{0.69}} & \num{0.30} \\
\bottomrule
\end{tabular}
\scriptsize
\begin{flushleft}
\textbf{Critical FN/FP}: FN tracks with MDR\,$\geq$\,\SI{4.0}{\meter\per\second\squared} and FP tracks with FSR\,$\geq$\,\SI{2.5}{\meter\per\second}. \textbf{TC}: time-critical tracks ($\mathrm{TTC_{RSB}} < \num{2}$\,\si{\second}). \textbf{Cum.\ Effort}: sum of per-track MDR / FSR / LEA across all error tracks. MDR in \si{\meter\per\second\squared}. FSR in \si{\meter\per\second}. LEA ($a_\perp$) in \si{\meter\per\second\squared}. Mod.: Modality (L\,=\,LiDAR, L+C\,=\,LiDAR\,+\,Camera).
\end{flushleft}
\vspace{-0.5cm}
\end{table*}

\section{Experimental Evaluation}
\label{sec:experimental_evaluation}

The main focus of our work is an effort-based criticality framework that translates 3D perception errors into the collision-avoidance effort they impose on the ego vehicle. We present experiments on two large-scale autonomous-driving datasets to validate that FSR and MDR capture safety-relevant severity information that conventional detection metrics, such as TTC and DST, miss.

\subsection{Experimental Setup and Implementation Details}
\label{sec:implementation_details}

Table~\ref{tab:reach_params} summarizes all the parameters. For collision filtering, we adopt moderate acceleration bounds $a_{\text{lon}} \in [-3.0, 2.0]$\,\si{\meter\per\second\squared} and $a_{\text{lat}} \leq 2.0$\,\si{\meter\per\second\squared}, in accordance with~\cite{schneider2021towards} who derived $1.88$\,\si{\meter\per\second\squared} from the 98th percentile of naturalistic driving data.
FSR and MDR are capped at $a_{\text{brake,max}} = 10.0$\,\si{\meter\per\second\squared} (emergency braking, $\mu \approx 1.0$)~\cite{gassmann2019towards}. LEA at $a_{\text{lat,cap}} = 5.0$\,\si{\meter\per\second\squared} ($\approx 0.5g$, rollover stability)~\cite{young2022rollover}. The reaction time $t_{\text{react}} = 0.3$\,\si{\second} represents end-to-end AV latency~\cite{khalil2024plm}. We use $T_{\text{horizon}} = 5.0$\,\si{\second} with $\Delta t = 0.1$\,\si{\second} (vs.\ $2.0$\,\si{\second} and $0.4$\,\si{\second} in~\cite{schneider2021towards}). The longer horizon captures distant misses and the finer resolution improves collision-time accuracy to $\pm 0.1$\,\si{\second}.
We evaluate on nuScenes~\cite{Fong2025nuScenesRP}, comprising \num{1000} driving scenes with 3D annotations at \num{2}\si{\hertz}, and Argoverse~\cite{Wilson2023Argoverse2N} with annotations at \num{10}\si{\hertz}. We evaluate three pipelines:\looseness=-1
\begin{itemize}
\item \textit{Megvii + AB3DMOT}~\cite{9341164}: 3D bounding boxes without velocity. AB3DMOT employs a Kalman filter to estimate velocity under a constant-velocity model.
\item \textit{CenterPoint}~\cite{9578166}: LiDAR-based detector that directly predicts velocity $[v_x, v_y]$ via a dedicated regression head.
\item \textit{BEVFusion}~\cite{10160968}: Multi-modal (LiDAR + Camera) detector that fuses modalities in bird's-eye-view space and predicts velocity through a CenterPoint-style head~\cite{9578166}.
\end{itemize}


Velocity estimates are obtained from AB3DMOT's Kalman filter or the CenterPoint-style velocity head. FN longitudinal acceleration is estimated per track using a time-normalized backward difference of successive annotated longitudinal velocities, clipped to $[-10,10]$\,\si{\meter\per\second\squared}. The first sample and invalid time intervals are set to zero acceleration. Values entering the solver are clipped by the vehicle limits in Tab.~\ref{tab:reach_params}. Predictions are matched to GT per frame via Hungarian assignment with a $2.0$\,\si{\meter} center-distance matching threshold~\cite{Fong2025nuScenesRP, Wilson2023Argoverse2N}. Unmatched ground-truth objects are labeled as FNs, and unmatched predictions as FPs. Errors are aggregated by object identity: MDR is the maximum deceleration in a FN track, whereas FSR integrates over the FP lifetime. Re-detected phantoms under new IDs are separate tracks.

\textit{Criticality Thresholds:}
Track-level errors are classified by operational severity into four zones (Safe, Moderate, Critical, Imminent) as summarized in Tab.~\ref{tab:criticality_thresholds}.
The MDR boundaries follow the deceleration scale used for DST~\cite{Westhofen_2022}: $\leq 2.0$\,\si{\meter\per\second\squared} (comfortable) through $> 6.0$\,\si{\meter\per\second\squared} (emergency, near the physical limit~\cite{gassmann2019towards}).
For FSR, the "imminent" threshold of $5.0$\,\si{\meter\per\second} ($= 18$\,\si{\kilo\meter\per\hour}) exceeds the Euro~NCAP AEB false-activation criterion ($\sim 3.5$\,\si{\meter\per\second})~\cite{euro2015assessment}. LEA thresholds span comfortable lane changes ($\leq 1.0$\,\si{\meter\per\second\squared}) to rollover stability limits ($> 4.0$\,\si{\meter\per\second\squared})~\cite{young2022rollover}.

\begin{table}[t]
\centering
\caption{Criticality Metric Thresholds for Perception Error Assessment}
\label{tab:criticality_thresholds}
\scriptsize
\setlength{\tabcolsep}{2.5pt}
\begin{tabular}{@{}l cc cc >{\columncolor{colhighlight}}c >{\columncolor{colhighlight}}c >{\columncolor{colhighlight}}c @{}}
\toprule
 & \multicolumn{2}{c}{\textit{Time-Based}} & \multicolumn{2}{c}{\textit{Decel.-Based}} & \multicolumn{3}{c}{\cellcolor{colhighlight}\textit{Effort-Based (Proposed)}} \\
\cmidrule(lr){2-3} \cmidrule(lr){4-5} \cmidrule(lr){6-8}
\textbf{Severity}
  & \textbf{TTC} & \textbf{TTB}
  & \textbf{BTN} & \textbf{DST}
  & \textbf{MDR} & \textbf{FSR} & \textbf{LEA} \\
 & (\si{\second}) & (\si{\second})
  & (-) & (\si{\meter/\second\squared})
  & (\si{\meter/\second\squared}) & (\si{\meter/\second}) & (\si{\meter/\second\squared}) \\
\midrule
Safe     & ${>}3.0$ & ${>}1.0$ & ${\leq}0.4$ & ${<}1.0$ & ${\leq}2.0$ & ${\leq}1.0$ & ${\leq}1.0$ \\
Moderate & $2.0$-$3.0$ & - & $0.4$-$0.7$ & $1.0$-$4.0$ & $2.0$-$4.0$ & $1.0$-$2.5$ & $1.0$-$2.0$ \\
Critical & $1.0$-$2.0$ & $0.4$-$1.0$ & $0.7$-$1.0$ & $4.0$-$6.0$ & $4.0$-$6.0$ & $2.5$-$5.0$ & $2.0$-$4.0$ \\
Imminent & ${<}1.0$ & ${<}0.4$ & ${>}1.0$ & ${\geq}6.0$ & ${>}6.0$ & ${>}5.0$ & ${>}4.0$ \\
\bottomrule
\end{tabular}
\vspace{1mm}
\begin{flushleft}
\scriptsize
\textit{Thresholds:} TTC critical at $1$-$2$\,\si{\second}~\cite{junietz2018criticality}. TTB MRM threshold: $0.4$\,\si{\second}~\cite{Westhofen_2022}. FSR of $3.5$\,\si{\meter\per\second} $\approx$ $12.6$\,\si{\kilo\meter\per\hour} reduction in traffic disruption~\cite{euro2015assessment}. MDR derived from driving guidelines ($a_{\text{brake}} \in [4, 8]$\,\si{\meter\per\second\squared})~\cite{gassmann2019towards}. Physical limits: Refer to Table~\ref{tab:reach_params}.
\end{flushleft}
\vspace{-0.4cm}
\end{table}

\subsection{Quantitative Comparison}
\label{sec:main_results}

Table~\ref{tab:comprehensive_results} presents the effort-based safety evaluation across three perception pipelines on both datasets. Beyond mean and worst-case effort per track, we report \emph{cumulative effort} as the sum of MDR or FSR across all track errors. Cumulative effort captures both the severity and volume of errors.

\subsubsection{Cumulative effort disambiguates pipelines}
On nuScenes, CenterPoint yields more FP tracks than AB3DMOT (\num{40172} vs \num{36239} for cars) with lower mean FSR (\num{0.99} vs \num{1.66}\si{\meter\per\second}), so standard precision alone cannot determine which pipeline imposes a greater operational burden. Cumulative FSR resolves this. AB3DMOT accumulates \num{60242}\si{\meter\per\second} of unnecessary braking compared to \num{39955}\si{\meter\per\second} for CenterPoint, confirming that AB3DMOT's fewer but more persistent phantoms are more costly overall. BEVFusion reduces cumulative FSR by an order of magnitude to \num{4723}\si{\meter\per\second} for cars, the clearest indicator of its FP advantage.

\begin{figure}[t]
\centering
\includegraphics[width=\columnwidth]{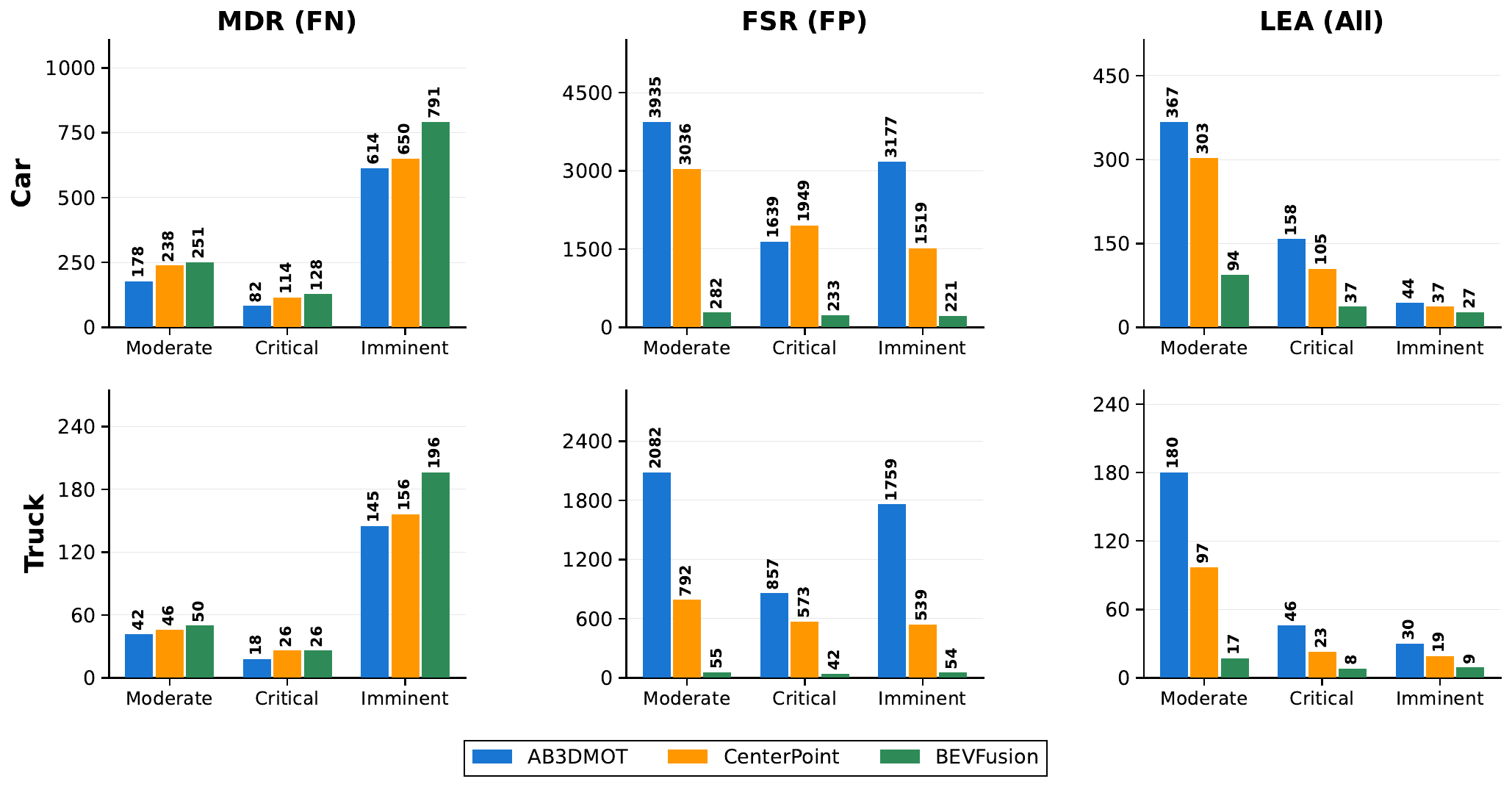}
\caption{Distribution of non-safe perception errors (absolute track counts) on the nuScenes validation set. Rows: Car / Truck. Columns: FN by MDR, FP by FSR, all by LEA. Severity thresholds follow Tab.~\ref{tab:criticality_thresholds}.}
\label{fig:criticality_distribution}
\vspace{-0.4cm}
\end{figure}

\subsubsection{Cumulative MDR discriminates FN risk}
Mean MDR alone is not discriminative: it varies only from \num{2.01} to \num{2.53}\si{\meter\per\second\squared} for cars, and the worst event reaches the \num{10.0}\si{\meter\per\second\squared} cap in every configuration. These statistics describe the per-track severity distribution, which is dominated by scenario geometry. The discriminative quantity is cumulative MDR, which combines FN frequency with per-track severity. On nuScenes cars it rises from \num{7550} (AB3DMOT) to \num{9766} (BEVFusion), and on Argoverse~2 cars from \num{16419} (CenterPoint) to \num{21214} (AB3DMOT). Cumulative MDR therefore separates pipelines whose mean severities are nearly identical, at the evaluated operating point.

\subsubsection{Cross-dataset observations}
By Eq.~\ref{eq:fsr_cumulative}, FSR depends on the error duration $N_{\text{frames}}\times\,T_{\text{cycle}}$, not on the annotation frequency itself. The lower mean FSR on Argoverse~2 therefore reflects the much larger number of short-lived FP tracks surfaced at \num{10}\si{\hertz}, together with different scene composition, rather than a rescaling by the time step. The cross-pipeline ordering is only partially consistent across datasets: BEVFusion attains the lowest cumulative FSR and LEA on both, whereas AB3DMOT and CenterPoint exchange ranks between nuScenes and Argoverse~2 (cumulative FSR of \num{60242} vs \num{39955} on nuScenes, but \num{5315} vs \num{12025} on Argoverse~2 for cars).

\begin{table}[t]
\centering
\caption{Established vs proposed effort-based criticality metrics for assessing perception error.}
\label{tab:metric_properties}
\scriptsize
\setlength{\tabcolsep}{2.5pt}
\begin{tabular}{@{}l ccccc ccc@{}}
\toprule
 & \multicolumn{5}{c}{\textit{Established}} & \multicolumn{3}{c}{\cellcolor{colhighlight}\textit{Proposed}} \\
\cmidrule(lr){2-6} \cmidrule(lr){7-9}
\textbf{Property} & \textbf{TTC} & \textbf{DRAC} & \textbf{BTN} & \textbf{DST} & \textbf{TET} & \cellcolor{colhighlight}\textbf{MDR} & \cellcolor{colhighlight}\textbf{FSR} & \cellcolor{colhighlight}\textbf{LEA} \\
\midrule
FP/FN specific        & -   & -   & -   & -   & -   & \cellcolor{colhighlight}FN  & \cellcolor{colhighlight}FP  & \cellcolor{colhighlight}Both \\
Temporal scope         & inst.  & inst.  & inst.  & inst.  & cum. & \cellcolor{colhighlight}peak & \cellcolor{colhighlight}cum. & \cellcolor{colhighlight}peak \\
Duration-sensitive     & $\times$ & $\times$ & $\times$ & $\times$ & $\checkmark$ & \cellcolor{colhighlight}$\times$ & \cellcolor{colhighlight}$\checkmark$ & \cellcolor{colhighlight}$\times$ \\
Lateral dimension      & $\times$ & $\times$ & $\times$ & $\times$ & $\times$ & \cellcolor{colhighlight}$\times$ & \cellcolor{colhighlight}$\times$ & \cellcolor{colhighlight}$\checkmark$ \\
Collision gate         & -   & -   & -   & -   & -   & \cellcolor{colhighlight}RSB  & \cellcolor{colhighlight}RSB  & \cellcolor{colhighlight}RSB \\
Kinematic model        & CV   & CV   & CV   & CV   & CV   & \cellcolor{colhighlight}CA & \cellcolor{colhighlight}CV   & \cellcolor{colhighlight}CV  \\
Output unit            & \si{\second} & \si{\meter\per\second\squared} & - & \si{\meter\per\second\squared} & \si{\second} & \cellcolor{colhighlight}\si{\meter\per\second\squared} & \cellcolor{colhighlight}\si{\meter\per\second} & \cellcolor{colhighlight}\si{\meter\per\second\squared} \\
\midrule
Equiv.\ in~\cite{Westhofen_2022} & TTC & $a_{\text{long,req}}$ & BTN & DST & TET & \cellcolor{colhighlight}\makecell{\textit{novel}} & \cellcolor{colhighlight}\makecell{\textit{novel}} & \cellcolor{colhighlight}$a_{\text{lat,req}}$ \\
\bottomrule
\end{tabular}
\begin{flushleft}
\scriptsize inst.\,=\,instantaneous, cum.\,=\,cumulative, peak\,=\,per-frame maximum, CV\,=\,constant velocity, CA\,=\,constant acceleration (Eq.~\ref{eq:mdr_base}). DRAC is the CV special case of $a_{\text{long,req}}$ ($a_{\text{obj}}{=}0$)~\cite{Westhofen_2022}, DST~\cite{Westhofen_2022} extends DRAC with a safety-time buffer. RSB\,=\,reachable-set-based collision filter (Sec.~\ref{sec:reachability})~\cite{schneider2021towards}, SAT\,=\,OBB overlap~\cite{10.1145/3727353.3727481}.
\end{flushleft}
\vspace{-0.4cm}
\end{table}


\subsection{Error Severity Distribution}
\label{sec:scatter_analysis}

Fig.~\ref{fig:criticality_distribution} shows the distribution of \emph{non-safe} tracks on the nuScenes validation set, classified into Moderate, Critical, and Imminent severity zones defined in Tab.~\ref{tab:criticality_thresholds}. The Safe zone, which accounts for \SIrange{70}{84}{\percent} of FN tracks by MDR, \SIrange{76}{93}{\percent} of FP tracks by FSR, and \SIrange{98}{99}{\percent} of all tracks by LEA, is omitted to expose the safety-relevant tail, confirming that raw error counts substantially overestimate actual risk. Every pipeline yields several hundred Imminent FN tracks, but these counts are upper bounds set by the conservative reachable-set gate rather than verified emergency-braking cases. Because the gate encodes only bounded motion and no lane or map semantics, laterally distant objects, such as oncoming vehicles in adjacent lanes, can be admitted and scored with high braking demand. On the nuScenes car category, roughly \SI{90}{\percent} of MDR-critical FN tracks lie off the ego path by the lateral-offset proximity defined as $\min|d_y|>1.75$\,\si{\meter}, with median lateral offsets above \SI{20}{\meter}. Cross-pipeline comparisons remain informative because all pipelines share the same collision gate. BEVFusion's Imminent FN fraction is the highest, suggesting that easier-to-detect objects are found first while the most challenging cases remain. On the FP side, BEVFusion's LiDAR-camera fusion suppresses phantom detections, yielding far fewer Imminent FPs. LEA is predominantly safe, with fewer than \num{300} tracks per pipeline in the Critical or Imminent zones, suggesting that most errors involve laterally offset objects that require minimal steering effort.


\subsection{Proposed Metrics vs.\ Established Criticality Metrics}
\label{sec:criticality_distribution}

We now examine whether the effort metrics provide rankings complementary to established criticality measures. Table~\ref{tab:metric_properties} highlights three structural distinctions: (i)~FSR and MDR are error-type specific, (ii)~FSR is duration-sensitive, and (iii)~LEA adds a lateral dimension. To quantify these differences, we extend the comparison with THW\,($R/v_{\text{ego}}$), Time Exposed TTC-threshold (TET, cumulative time with TTC$<2$\,\si{\second})~\cite{Westhofen_2022}, lateral distance $d_y$, and compute track-level Spearman correlations ($\rho$) across the nuScenes pipelines under RSB and SAT collision checks in Tab.~\ref{tab:correlation}.

\begin{table}[t]
\centering
\caption{Spearman correlation ($\rho$) between the proposed and established metrics on nuScenes, aggregated across all pipelines under both collision checks. \textbf{Bold:} $|\rho| \geq 0.40$.}
\label{tab:correlation}
\scriptsize
\resizebox{\columnwidth}{!}{%
\setlength{\tabcolsep}{2pt}
\begin{tabular}{@{}ll rr rr rr rr rr@{}}
\toprule
 & & \multicolumn{2}{c}{\textbf{TTC}} & \multicolumn{2}{c}{\textbf{DRAC}} & \multicolumn{2}{c}{\textbf{THW}} & \multicolumn{2}{c}{\textbf{TET}} & \multicolumn{2}{c}{$\boldsymbol{d_y}$} \\
\cmidrule(lr){3-4} \cmidrule(lr){5-6} \cmidrule(lr){7-8} \cmidrule(lr){9-10} \cmidrule(lr){11-12}
 & & \cellcolor{colhighlight}RSB & SAT & \cellcolor{colhighlight}RSB & SAT & \cellcolor{colhighlight}RSB & SAT & \cellcolor{colhighlight}RSB & SAT & \cellcolor{colhighlight}RSB & SAT \\
\midrule
\multirow{2}{*}{FN}
 & MDR & \cellcolor{colhighlight}$\text{-}0.30$ & $\text{-}0.34$ & \cellcolor{colhighlight}$0.30$ & $\mathbf{0.68}$ & \cellcolor{colhighlight}$\text{-}0.39$ & $0.07$ & \cellcolor{colhighlight}$\mathbf{0.41}$ & $0.15$ & \cellcolor{colhighlight}$0.11$ & $0.16$ \\
 & LEA & \cellcolor{colhighlight}$0.14$ & $0.31$ & \cellcolor{colhighlight}$\text{-}0.12$ & $\text{-}0.36$ & \cellcolor{colhighlight}$0.10$ & $0.16$ & \cellcolor{colhighlight}$\text{-}0.13$ & $\text{-}0.10$ & \cellcolor{colhighlight}$\mathbf{\text{-}0.55}$ & $\mathbf{\text{-}0.58}$ \\
\midrule
\multirow{2}{*}{FP}
 & FSR & \cellcolor{colhighlight}$\mathbf{\text{-}0.93}$ & $\mathbf{\text{-}0.65}$ & \cellcolor{colhighlight}$\mathbf{0.89}$ & $\mathbf{0.80}$ & \cellcolor{colhighlight}$\mathbf{\text{-}0.90}$ & $\text{-}0.25$ & \cellcolor{colhighlight}$\mathbf{0.73}$ & $0.31$ & \cellcolor{colhighlight}$0.06$ & $0.11$ \\
 & LEA & \cellcolor{colhighlight}$0.08$ & $\text{-}0.02$ & \cellcolor{colhighlight}$\text{-}0.07$ & $0.02$ & \cellcolor{colhighlight}$0.06$ & \reviewnew{$0.04$} & \cellcolor{colhighlight}$\text{-}0.08$ & $\text{-}0.06$ & \cellcolor{colhighlight}$\mathbf{\text{-}0.41}$ & $\mathbf{\text{-}0.57}$ \\
\bottomrule
\end{tabular}%
}
\begin{flushleft}
\scriptsize RSB\,=\,reachable-set-based collision filter (Sec.~\ref{sec:reachability})~\cite{schneider2021towards}, SAT\,=\,OBB overlap~\cite{10.1145/3727353.3727481}. All pairwise $|\rho|$ among established metrics exceed $0.69$ (RSB). $d_y$\,=\,min.\ lateral distance. \textit{Scored events} (RSB\,/\,SAT): FN\,=\,\num{12446}\,/\,\num{303}, FP\,=\,\num{107937}\,/\,\num{1465}, \textit{Crit.\,/\,Imm.}: FN\,=\,\num{2946}\,/\,\num{47}, FP\,=\,\num{12562}\,/\,\num{99}.
\end{flushleft}
\vspace{-0.7cm}
\end{table}

\begin{figure*}[t]
\centering
\includegraphics[width=\textwidth]{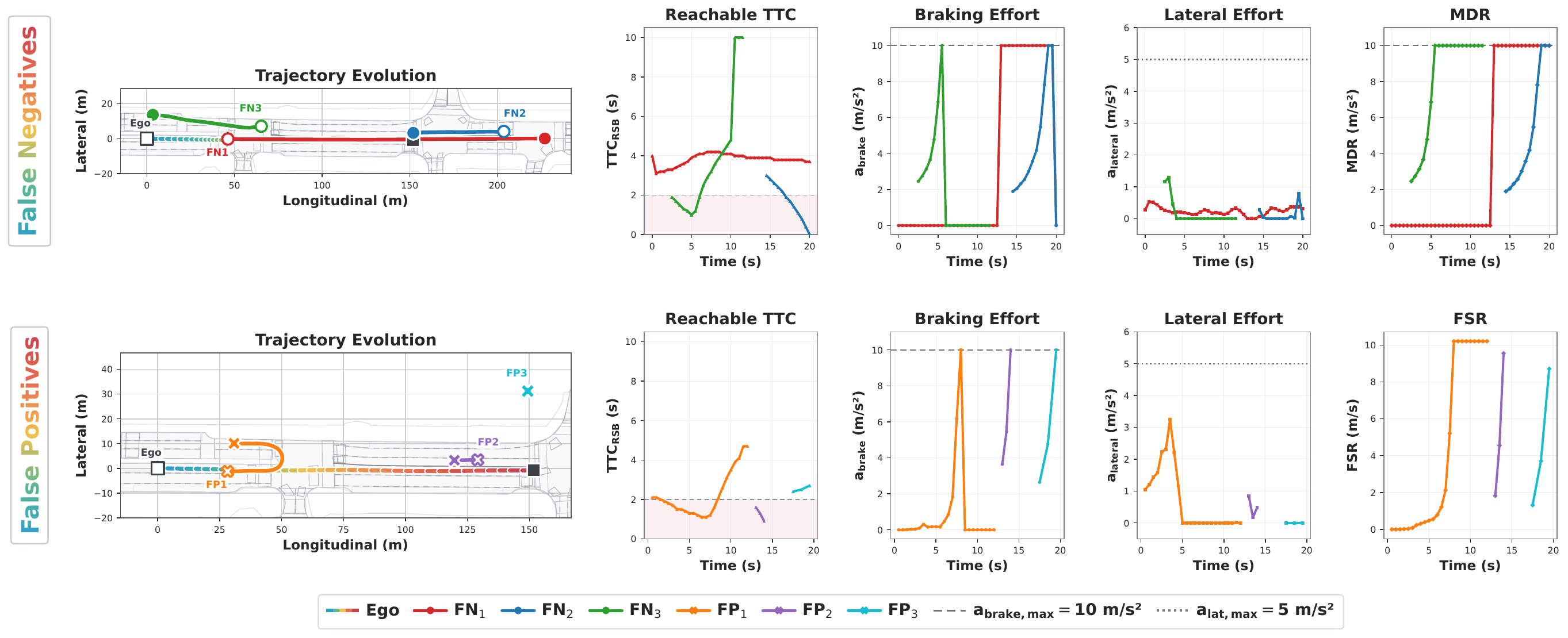}
\caption{%
\textbf{Scenario-level analysis} for scene \texttt{26a6b03c} (nuScenes, BEVFusion). Each row shows, from left to right: bird's-eye trajectory evolution with start/end markers, reachability-based TTC ($\text{TTC}_{\text{RSB}}$), braking effort ($a_{\text{brake}}$), lateral evasion effort (LEA), and the cumulative criticality metric (MDR for FNs, FSR for FPs). \textit{Top}: three FN objects ($\text{FN}_{1\text{-}3}$) illustrate how MDR saturates for both in-path and laterally distant objects, while LEA disambiguates true collision risk. \textit{Bottom}: three FP objects ($\text{FP}_{1\text{-}3}$) demonstrate FSR's persistence-sensitive accumulation. $\text{FP}_1$ (24 frames, low per-frame braking) scores highest, while short-lived FPs with higher intensity receive lower FSR.
}
\label{fig:scenario_analysis}
\vspace{-0.5cm}
\end{figure*}

\subsubsection{MDR captures unique information}

Under RSB gating, MDR correlates weakly with the established metrics, with correlation magnitudes of at most $0.41$. Allowing object acceleration in Eq.~\ref{eq:mdr_base} can rank closing interactions differently from CV-based DRAC, where $\rho=0.30$, while the broad set admitted by RSB also changes the evaluated population. Under SAT, $\rho(\text{MDR},\text{DRAC})$ rises to $0.68$. Among \num{3237} FN tracks with $\text{MDR}>3$\,\si{\meter\per\second\squared}, \SI{42}{\percent} have $\text{TET}=0$ and \SI{32}{\percent} have TTC above $4$\,\si{\second}. These tracks will not be selected by the corresponding TTC thresholds.

\subsubsection{LEA has weak rank association with longitudinal metrics}
LEA remains weakly correlated with TTC, DRAC, THW, and TET under RSB, with correlation magnitudes below $0.15$. This supports complementarity on the evaluated sample, rather than statistical or causal independence. Correlation with lateral distance $d_y$ is $-0.55$ (FN) and $-0.41$ (FP) under RSB, and $-0.58$ and $-0.57$ under SAT. The gap to unity indicates that LEA reflects not only offset but also evasion time and relative lateral velocity.

\subsubsection{FSR adds temporal depth to DRAC}
FSR shares the CV model with DRAC and, accordingly, correlates with DRAC at $\rho = 0.89$ for FP tracks. The unique contribution is duration-sensitivity: \SI{17}{\percent} of \num{14471} tracks with $\text{FSR} > 2$\,\si{\meter\per\second} have $\text{TET} = 0$, meaning these persistent phantoms never individually breach $\text{TTC} < 2$\,\si{\second} yet accumulate substantial unnecessary braking.

\subsubsection{Established metrics are mutually redundant}
\begin{figure}[t]
\centering
\begin{subfigure}{\columnwidth}
\centering
\includegraphics[width=\columnwidth]{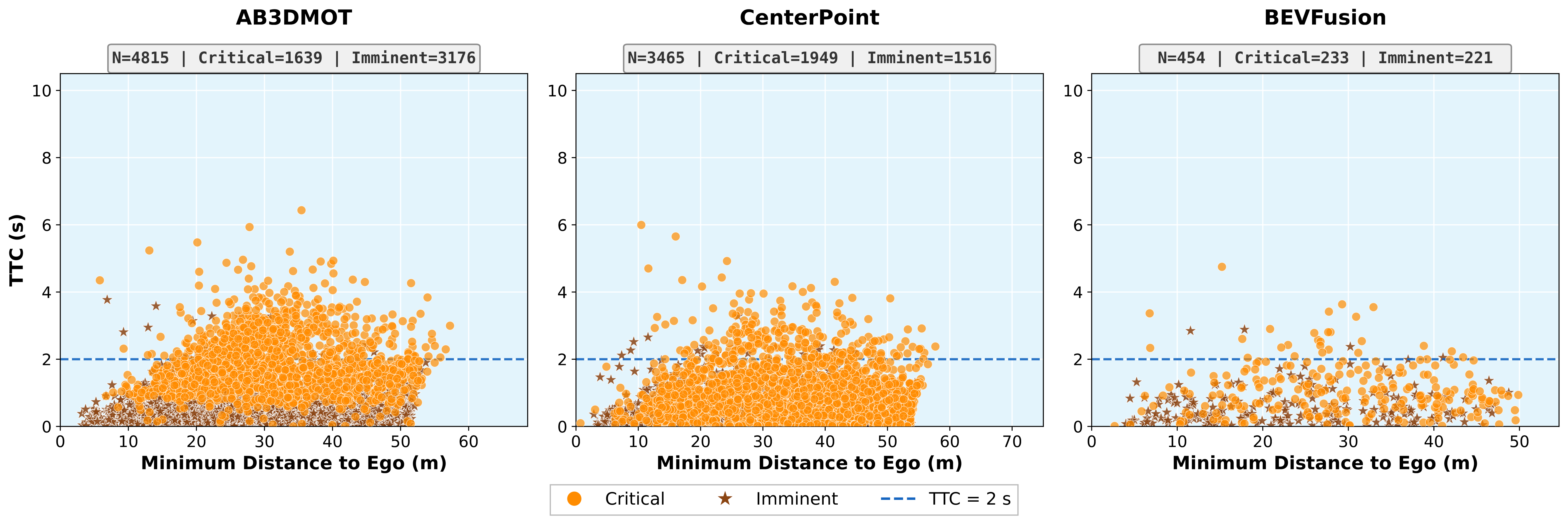}
\caption{Critical/Imminent FP tracks (FSR thresholds): TTC vs.\ minimum distance to ego.}
\label{fig:scatter_fsr}
\end{subfigure}
\vspace{0.6em}
\begin{subfigure}{\columnwidth}
\centering
\includegraphics[width=\columnwidth]{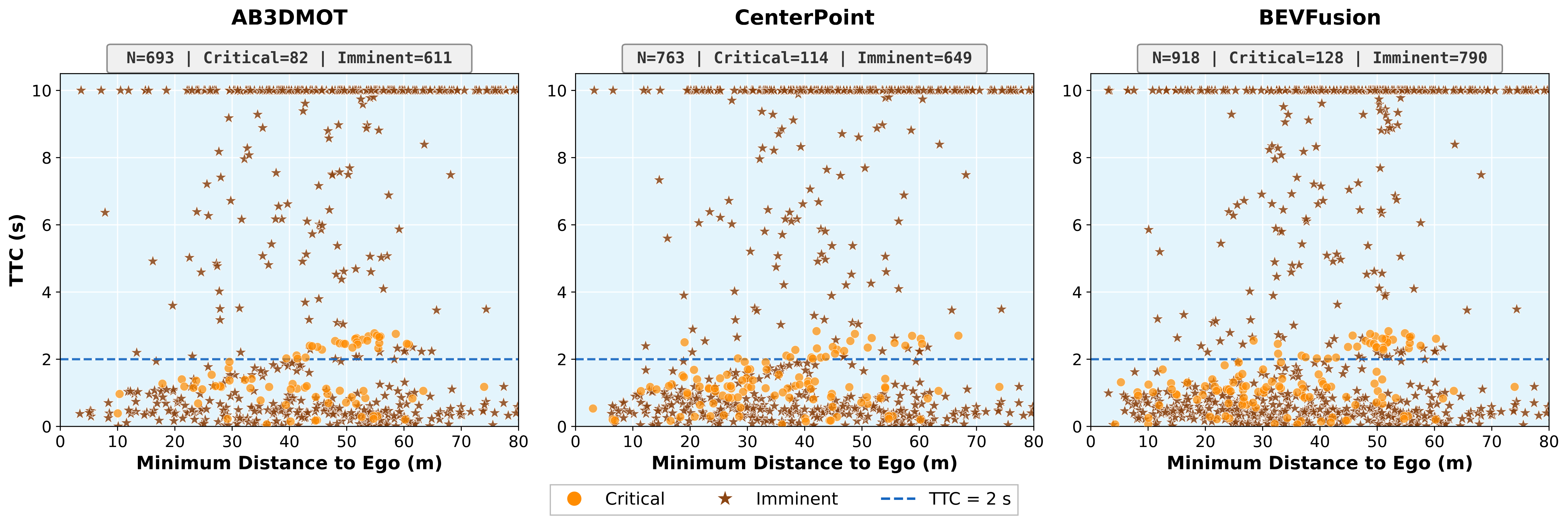}
\caption{Critical/Imminent FN tracks (MDR thresholds): TTC vs.\ minimum distance to ego.}
\label{fig:scatter_mdr}
\end{subfigure}
\caption{Classical TTC ($d/v_{\text{rel}}$) vs.\ minimum distance for effort-critical car-category tracks on the nuScenes validation set. Dashed line: TTC$=2$\,\si{\second}. A substantial fraction of MDR-critical FN tracks cluster at TTC${>}3$\,\si{\second} or at the dataset cap of $10$\,\si{\second}, indicating that classical TTC rates them as safe despite high braking demand.}
\label{fig:scatter_car_both}
\vspace{-0.75cm}
\end{figure}

All pairwise correlation magnitudes among established criticality metrics exceed $0.69$ under RSB, indicating that any single CV-based measure largely subsumes the others. Figs.~\ref{fig:scatter_fsr} and~\ref{fig:scatter_mdr} confirm this spatially: FSR-critical FP tracks cluster below $\text{TTC} = 2$\,\si{\second}, whereas MDR-critical FN tracks span $5$-$80$\,\si{\meter} with many above $\text{TTC} = 3$\,\si{\second}, a population that no established metric flags.

\subsection{Scenario-Level Analysis}
\label{sec:scenario_analysis}

We analyze three FNs and three FPs from scene \texttt{26a6b03c} of the nuScenes val set in Fig.~\ref{fig:scenario_analysis} using the rule-agnostic RSB collision check as explained in Sec.~\ref{sec:reachability}. As RSB does not encode lane structure, opposite-road or off-lane objects can be scored if dynamically reachable.

\subsubsection{False Negative Analysis}
$\text{FN}_1$ ($d_y \approx 0.7$\,\si{\meter}) is an in-lane lead vehicle missed for 38 frames. MDR reaches the cap of $10$\,\si{\meter\per\second\squared}, yet LEA is only $\approx 0.5$\,\si{\meter\per\second\squared}$-$a gentle lane change suffices. Reporting both metrics jointly reveals this braking-vs-steering trade-off.
$\text{FN}_2$ ($d_y \approx 4.3$\,\si{\meter}) and $\text{FN}_3$ ($d_y \approx 10.6$\,\si{\meter}) are on the opposite road, flagged only because RSB's reachable sets overlap under bounded lateral acceleration. All three FNs remain within the safe zone w.r.t. LEA given by Tab.~\ref{tab:criticality_thresholds} despite elevated MDR ($10$\,\si{\meter\per\second\squared}).

\subsubsection{False Positive Analysis}
$\text{FP}_1$ persists for $24$ frames with moderate per-frame braking ($1.46$\,\si{\meter\per\second\squared}), yet FSR assigns the highest score ($17.5$\,\si{\meter\per\second}) because disruption compounds over time. $\text{FP}_2$ and $\text{FP}_3$ last only $3$ frames with higher intensity of $6.4$ and $5.8$\,\si{\meter\per\second\squared} yet receive lower FSR of $9.5$ and $8.7$\,\si{\meter\per\second} respectively, confirming persistence over intensity. $\text{FP}_2$ ($d_y \approx 4.6$\,\si{\meter}, opposite road) and $\text{FP}_3$ ($d_y \approx 32$\,\si{\meter}, off-roadway) are scored only by RSB's conservative overlap. LEA correctly disambiguates: for both, the predicted lateral offset already clears the safety corridor, so no active steering is required ($\text{LEA}=0$). Under SAT, neither would pass the collision check.


This scene illustrates the intended screening workflow: MDR and FSR rank braking-demand candidates, while LEA and a tighter or map-aware gate are needed to reject laterally irrelevant cases. No individual score establishes that an event is safety-critical.


\section{Discussion}
\label{sec:discussion}

\subsubsection{Influence of matching on criticality}
FP/FN labels depend on the matching criterion. With a $2.0$\,\si{\meter} center-distance threshold, an object offset by $2.1$\,\si{\meter} is labelled an FN and may receive a high MDR even though a real planner could accommodate it. The metrics are therefore heuristic approximations but provide substantially richer information than binary detection counts. Refining matching towards ego-centric or safety evaluation is a complementary direction~\cite{kaul2026contourerrorsegocentricmatching}.

\subsubsection{Combining lateral and longitudinal efforts}
FSR and MDR measure longitudinal effort, whereas LEA measures lateral demand. We report them separately to preserve the avoidance mode. Combining them with a normalized factor, such as $\text{STN}=\text{LEA}/a_{\text{lat,max}}$~\cite{Westhofen_2022}, could obscure laterally converging cases.

\subsubsection{Role of the collision filter} The RSB collision check is dynamically conservative and semantically unfiltered, admitting \SI{98}{\percent} of FN and \SI{93}{\percent} of FP tracks, whereas SAT retains only \SI{2}{\percent} and \SI{1}{\percent} given by Tab.~\ref{tab:correlation}. A map-aware occupancy or route-corridor test can replace either gate without changing the effort formulations.

\subsubsection{Benchmark and runtime use} For benchmark integration, detection/tracking outputs are matched once, the chosen collision gate is applied per frame, and FSR/MDR/LEA are aggregated per error track alongside AP and tracking metrics. All three metrics are offline diagnostics. They require ground-truth, FP/FN labels and lack the map, occupancy, and prediction context available to an onboard planner or AEB system, so they are neither computable nor actionable online.

\subsubsection{Toward objective evaluation}  All pipelines are evaluated at a single detection-score threshold of $0.1$, which maps to different precision-recall operating points because the detectors are calibrated differently. A matched-recall comparison or a cumulative effort vs recall curve would remove this dependence, but would require re-thresholding, rematching, and rebuilding tracks at every confidence value. We therefore report operating-point dependence as a limitation rather than claiming threshold-independent model superiority.


\section{Conclusion}
\label{sec:conclusion}

We formulated FSR and MDR as error-specific effort metrics for open-loop 3D perception evaluation: FSR accumulates FP-associated braking demand and MDR records peak FN-associated braking demand. LEA adds a lateral-evasion view. Across the evaluated operating points, \SIrange{65}{93}{\percent} of errors fall below the selected criticality thresholds. The collision gate check also exposes an important limitation: reachability is dynamically conservative but semantically unfiltered, so the resulting scores are suitable for candidate screening and failure mining, not safety certification. Future evaluation should add map-aware gating, manual or closed-loop validity labels, and confidence-sweep effort-recall curves.

\footnotesize
\bibliographystyle{IEEEtran}
\bibliography{references.bib}

\end{document}